\documentclass[10pt,twocolumn,letterpaper]{article}

\usepackage[pagenumbers]{cvpr} 

\usepackage[accsupp]{axessibility}
\usepackage{times}
\usepackage{epsfig}
\usepackage{graphicx}
\usepackage{amsmath}
\usepackage{amssymb}
\usepackage{multirow}
\usepackage{array}
\usepackage{enumitem}
\usepackage{booktabs}
\usepackage{subcaption}
\usepackage{refcount,xcolor}

\definecolor{citecolor}{HTML}{0071bc}
\usepackage[pagebackref,breaklinks=true,letterpaper=true,colorlinks,citecolor=citecolor,bookmarks=false]{hyperref}

\usepackage[capitalize]{cleveref}
\crefname{section}{Sec.}{Secs.}
\Crefname{section}{Section}{Sections}
\Crefname{table}{Table}{Tables}

\def\cvprPaperID{2029} 
\def\confName{CVPR}
\def\confYear{2022}

\newcommand*{\affaddr}[1]{#1} 
\newcommand*{\affmark}[1][*]{\textsuperscript{#1}}
\newcommand*{\email}[1]{\texttt{#1}}

\usepackage{cite,nicefrac}
\usepackage{listings}

\usepackage{enumitem}
\setitemize{noitemsep,topsep=3pt,parsep=0pt,partopsep=0pt}

\usepackage[margin=4pt,font=small,labelfont=bf,labelsep=endash,tableposition=top]{caption}

\begin{document}

\title{TopFormer: Token Pyramid Transformer for Mobile Semantic Segmentation}

\author{%
    Wenqiang Zhang\affmark[1]\thanks{Equal contribution. This work was done when W. Zhang was an intern at Tencent GY-Lab.}, Zilong Huang\affmark[2]\footnotemark[1], Guozhong Luo\affmark[2], Tao Chen\affmark[3], \\ Xinggang Wang\affmark[1]\thanks{Corresponding authors}, Wenyu Liu\affmark[1]\footnotemark[2], Gang Yu\affmark[2], Chunhua Shen\affmark[4] \\
    \affaddr{\affmark[1]{Huazhong University of Science
    and Technology}} 
    \affaddr{\affmark[2]{Tencent PCG} } 
    \affaddr{\affmark[3]{Fudan University}}
    \affaddr{\affmark[4]{Zhejiang University}} \\
    \email{\tt\small \{wq\_zhang,xgwang,liuwy\}@hust.edu.cn  \{zilonghuang,alexantaluo,skicyyu\}@tencent.com } \\
    \email{\tt\small eetchen@fudan.edu.cn chhshen@gmail.com}
    }
%
%
%
\maketitle

\begin{abstract}
    Although vision transformers (ViTs) have achieved great success in computer vision, the heavy computational cost hampers their applications to 
    dense prediction tasks such as semantic segmentation on mobile devices. In this paper, we present a mobile-friendly architecture named \textbf{To}ken \textbf{P}yramid Vision Trans\textbf{former} (\textbf{TopFormer}). The proposed \textbf{TopFormer} takes Tokens from various scales as input to produce scale-aware semantic features, which are then injected into the corresponding tokens to augment the representation. Experimental results demonstrate that our method significantly outperforms CNN- and ViT-based networks across several semantic segmentation datasets and achieves a good trade-off between accuracy and latency. On the ADE20K dataset, TopFormer achieves 5\% higher accuracy in mIoU than MobileNetV3 with lower latency on an ARM-based mobile device. Furthermore, the tiny version of TopFormer achieves real-time inference on an ARM-based mobile device with competitive results. The code and models are available at: 
    
    \url{https://github.com/hustvl/TopFormer}
    %
    %
    
\end{abstract}

\section{Introduction} \label{sec:intro}

Vision Transformers (ViTs) have 
shown considerably stronger results for a few vision tasks, 
such as image classification~\cite{dosovitskiy2020image}, object detection~\cite{liu2021swin}, and semantic segmentation~\cite{zheng2020rethinking}. Despite the success, the Transformer architecture with 
the 
full-attention mechanism \cite{vaswani2017attention} requires powerful computational resources beyond the capabilities of many mobile and embedded devices. In this paper, we aim to explore a mobile-friendly Vision Transformer specially designed for dense prediction tasks, \textit{e.g.}, semantic segmentation. 

\begin{figure}[!t]
    \centering
    \includegraphics[width=1\linewidth]{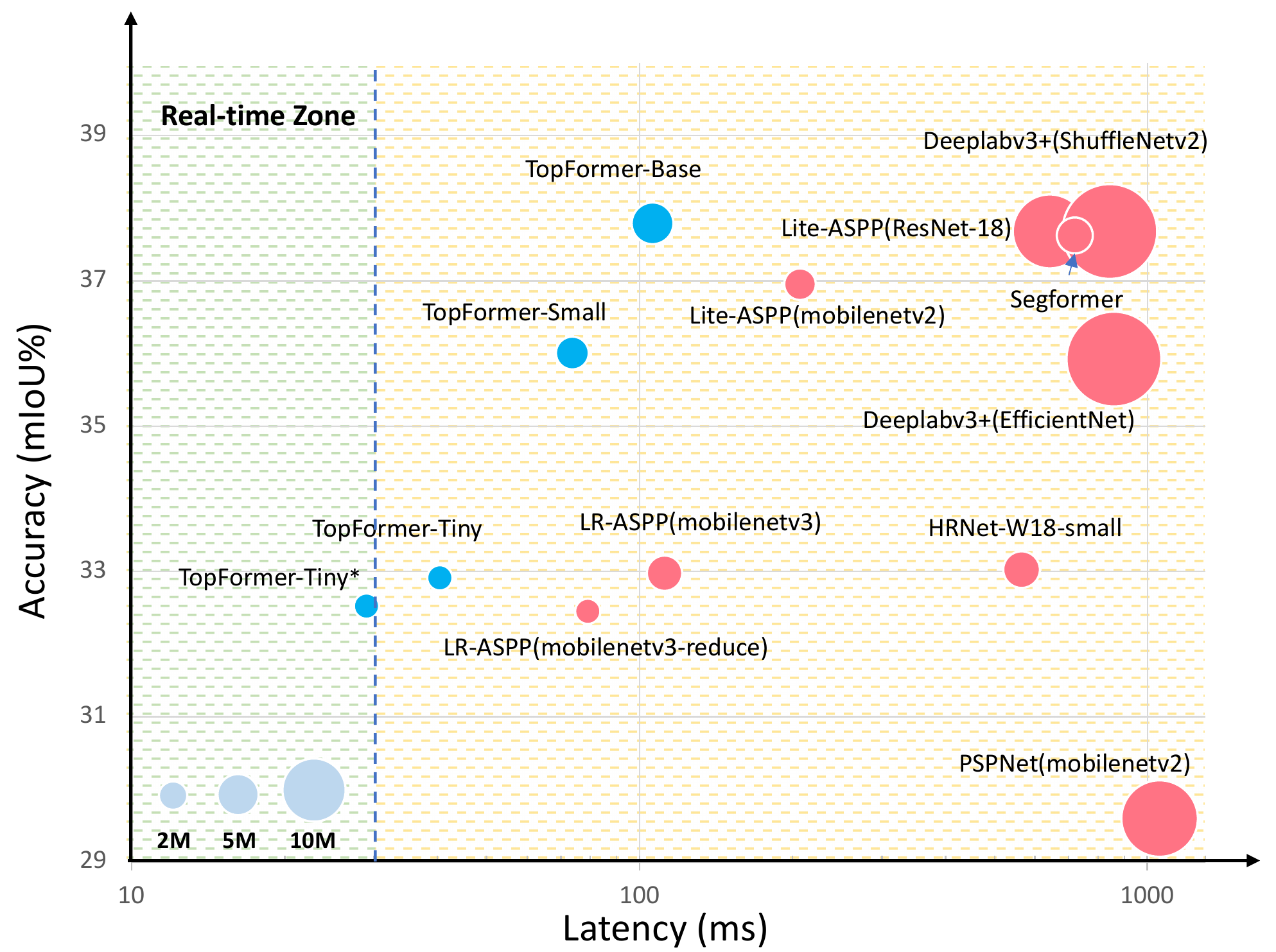}
    \caption{The latency,
    mIoU performance versus model size on
    the 
    ADE20K val.\  set. Previous
models are marked as red points, and our models are shown in blue points.
Our methods 
achieve
a better latency/accuracy trade-off. The latency is measured 
on a single Qualcomm Snapdragon 865 with input size 512$\times$512, and
only an ARM 
CPU
core is used for speed testing. No other means of acceleration, \textit{e.g.}, GPU or quantification, is used. * indicates the input size is 448$\times$448. } 
    \label{fig:speed_iou} 
    \vspace{-5mm}
\end{figure}

To adapt Vision Transformers to various dense prediction tasks, most 
recent Vision Transformers 
such as 
PVT~\cite{wang2021pyramid}, CvT~\cite{wu2021cvt}, LeViT~\cite{graham2021levit}, MobileViT~\cite{mehta2021mobilevit} adopt 
a hierarchical architecture, which is generally used in Convolution Neural Networks (CNNs), \textit{e.g.}, AlexNet~\cite{krizhevsky2012imagenet}, ResNet~\cite{he2016deep}. These Vision Transformers apply the global self-attention and its variants on the high-resolution tokens, which bring 
heavy 
computation cost due to the quadratic complexity 
in 
the number of tokens.

To improve the efficiency, some recent works, \textit{e.g.}, Swin Transformer \cite{liu2021swin}, Shuffle Transformer \cite{huang2021shuffle}, Twins~\cite{chu2021twins} and HR-Former \cite{yuan2021hrformer},  compute self-attention within the local/windowed region. However, the window partition is surprisingly time-consuming on mobile devices. Besides, Token slimming~\cite{tang2021patch} and Mobile-Former~\cite{chen2021mobile} decrease calculation capacity by reducing the number of tokens, but sacrifice their recognition accuracy.

Among these Vision Transformers, MobileViT~\cite{mehta2021mobilevit} and Mobile-Former~\cite{chen2021mobile} are specially designed for mobile devices. They both combine the strengths of CNNs and ViTs. For image classification, MobileViT achieves better performance than MobileNets~\cite{sandler2018mobilenetv2,howard2019searching} with a similar number of parameters. Mobile-Former achieves better performance than MobileNets with a fewer number of FLOPs. However, they 
do not show 
advantages in actual latency on mobile devices compared to MobileNets, as reported in \cite{mehta2021mobilevit}.
It raises a question: \textit{Is it possible to design mobile-friendly networks which could achieve better performance on mobile semantic segmentation tasks than MobileNets with lower latency?}

Inspired by MobileViT and Mobile-Former, we also make use of the advantages of CNNs and ViTs. A CNN-based module, denoted as Token Pyramid Module, is used to process high-resolution images to produce local features\footnote{%
We use `features' and `tokens' interchangeably here. 
} pyramid quickly. Considering the very limited computing power on mobile devices, here we use a few stacked light-weight MobileNetV2 blocks with a fast down-sampling strategy to build a token pyramid. To obtain rich semantics and large receptive field, the ViT-based module, denoted as Semantics Extractor, is adopted and takes the tokens as input. To further reduce the computational cost, the average pooling operator is used to reduce tokens to an extremely small number, \textit{e.g.}, $\nicefrac{1}{(64\times64)}$ of the input size. Different from ViT~\cite{dosovitskiy2020image}, T2T-ViT~\cite{Yuan_2021_ICCV} and LeViT~\cite{graham2021levit} use the last output of the embedding layer as input tokens, we pool the tokens from different scales (stages) into the very small numbers (resolution) and concatenate them 
along the 
channel dimension. Then the new tokens are fed into the Transformer blocks to produce global semantics. Due to the residual connections in the Transformer block, the learned semantics are related to scales of tokens, denoted as scale-aware global semantics. 

To obtain powerful hierarchical features for dense prediction tasks, scale-aware global semantics is split by the channels of tokens from different scales, then the scale-aware global semantics are fused with the corresponding tokens to augment the representation. The augmented tokens are used as the input of the segmentation head. 

To demonstrate the effectiveness of our approach, we conduct 
experiments on the challenging segmentation datasets: ADE20K~\cite{zhou2017scene}, Pascal Context~\cite{mottaghi2014role} and COCO-Stuff~\cite{caesar2018coco}. We examine the latency on 
 hardware, \textit{i.e.}, an off-the-shelf ARM-based computing core.  As shown in Figure~\ref{fig:speed_iou}, our approach obtains better results than MobileNets with lower latency. To demonstrate the generalization of our approach, we also conduct experiments of object detection on
the 
COCO~\cite{lin2014microsoft} dataset.
To summarize, our contributions 
are as follows. 
\begin{itemize}
    \item The proposed TopFormer takes tokens from different scales as input,
    and 
    pools the tokens to the very small numbers, 
    in order to obtain scale-aware semantics with very light computation cost.
    \item The proposed Semantics Injection Module 
    can 
    inject the scale-aware semantics into the corresponding tokens to build powerful hierarchical features, which is critical to dense prediction tasks.
    \item The proposed base model 
    can 
    achieve 5\% mIoU better than that of MobileNetV3, with lower latency on an ARM-based mobile device on the ADE20K dataset. The tiny version 
    can 
    perform real-time segmentation on an ARM-based mobile device, with competitive results. 
\end{itemize}


\section{Related Work} \label{sec:related}

In this section, we review recent approaches in terms of three aspects: 1) Light-weight Vision Transformer, 2) Efficient Convolutional Neural Networks, 3) Mobile Semantic Segmentation.

\subsection{Light-weight Vision Transformers}
There are many explorations~\cite{wang2018non, hu2019local, zhao2020exploring} for the 
use 
of transformer structures in image recognition. 
ViT~\cite{dosovitskiy2020image} is the first work to apply a pure transformer to image classification,
achieving 
state-of-the-art performance. 
Following that, DeiT~\cite{touvron2020deit} introduces token-based distillation to reduce the amount of data necessary for training the transformer. T2T-ViT~\cite{Yuan_2021_ICCV} structures the image to tokens by recursively aggregating neighboring tokens into one token to reduce tokens length. Swin Transformer~\cite{liu2021swin} computes self-attention within each local window, resulting in linear computational complexity in the number of input tokens. However, these Vision Transformers and the follow-ups 
are often of a large number of parameters and 
heavy computation complexity.

To build a light-weight Vision Transformer, LeViT~\cite{graham2021levit} designs a hybrid architecture that uses stacked standard convolution layers with stride-2 to reduce the number of tokens, then appends an improved Vision Transformer to extract semantics. For classification task, LeViT 
can 
significantly outperform EfficientNet on CPU. MobileViT~\cite{mehta2021mobilevit} adopts the same strategy and uses the MobilenetV2 block instead of the standard convolution layer for downsampling the feature maps. Mobile-Former takes parallel structure with a bidirectional bridge and leverages the advantage of both MobileNet and transformer. However, the MobileViT and other ViT-based networks are significantly slower 
than 
MobileNets~\cite{howard2017mobilenets,sandler2018mobilenetv2,howard2019searching} on mobile devices, as reported in ~\cite{mehta2021mobilevit}. For
the 
segmentation task, the input images are always 
of 
high-resolution. 
Thus it is even 
more 
challenging 
for ViT-based networks to 
execute 
faster than MobileNets.
In this paper, we aim to design a light-weight Vision Transformer which 
can 
outperform MobileNets with lower latency for the segmentation task.

\begin{figure*}[!t]
    \centering
    \includegraphics[width=1\linewidth]{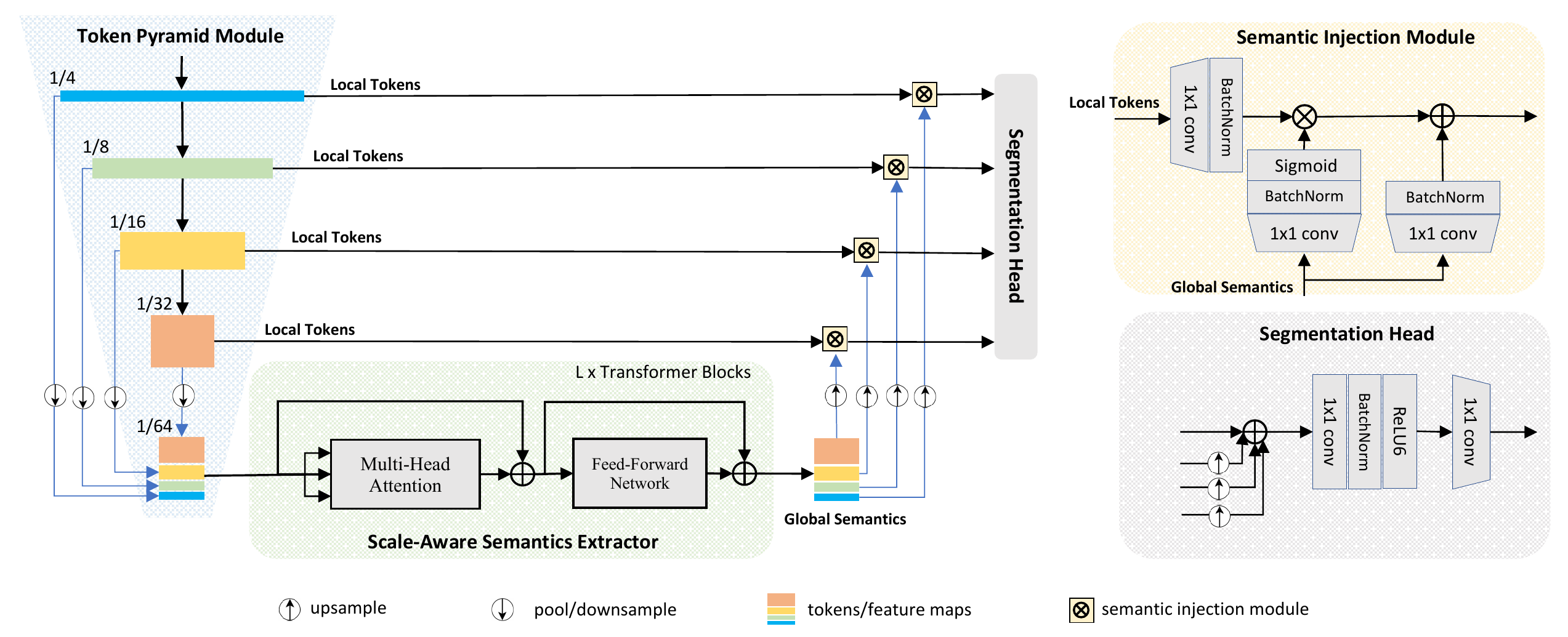}
    \caption{The architecture of the proposed Token Pyramid Transformer.} %
    \label{fig:architecture} 
    \vspace{-5mm}
\end{figure*}

\subsection{Efficient Convolutional Neural Networks}
The increasing need of 
deploying 
vision 
models 
on mobile and embedded devices encourages the study on efficient Convolutional Neural Networks designs.
MobileNet~\cite{howard2017mobilenets,sandler2018mobilenetv2,howard2019searching} proposes an inverted bottleneck structure which stacks depth-wise and point-wise
convolutions. IGCNet~\cite{zhang2017interleaved} and ShuffleNet~\cite{zhang2018shufflenet, ma2018shufflenet} use channel shuffle/permutation operators to the channel to make cross-group information flow for multiple group convolution layers. GhostNet~\cite{han2020ghostnet} uses the cheaper operator, depth-wise convolutions, to generate more features. AdderNet~\cite{chen2020addernet} utilizes additions to trade massive multiplications. MobileNeXt~\cite{zhou2020rethinking} flips the structure of the inverted residual block and presents a building block that connects high-dimensional representations instead. EfficientNet~\cite{tan2019efficientnet, tan2021efficientnetv2} and TinyNet~\cite{han2020model} study the compound scaling of depth, width and resolution.

\subsection{Mobile Semantic Segmentation}
The most accurate segmentation networks usually require computation at billions of FLOPs, which may exceed the computation capacity of the mobile and embedded devices.
To speed up the segmentation and reduce the computational cost, ICNet~\cite{zhao2018icnet} uses multi-scale images as input and a cascade network to be more efficient. DFANet~\cite{li2019dfanet} utilizes a light-weight backbone to speed up its network and proposes a cross-level feature aggregation to boost accuracy. SwiftNet~\cite{orsic2019defense} uses lateral connections as the cost-effective solution to restore the prediction resolution while maintaining the speed. BiSeNet~\cite{yu2018bisenet} introduces the spatial path and the semantic path to reduce computation. AlignSeg~\cite{huang2021alignseg} and SFNet~\cite{li2020semantic} align feature maps from adjacent levels and further enhances the feature maps using a feature pyramid framework.
ESPNets~\cite{mehta2019espnetv2} save computation by decomposing standard convolution into point-wise convolution and spatial pyramid of dilated convolutions. AutoML techniques~\cite{nekrasov2019fast,ding2021hrnas,chen2019fasterseg} are used to search 
for 
efficient architectures for scene parsing. NRD~\cite{zhang2021dynamic} dynamically generates the neural representations with dynamic convolution filter networks. LRASPP~\cite{howard2019searching} adopts MobileNetV3 as encoder and proposes a new efficient segmentation decoder Lite Reduced Atrous Spatial Pyramid Pooling (LR-ASPP), and it still serves as a strong baseline for mobile semantic segmentation.

\section{Architecture} \label{sec:arch}

Our  overall network architecture is shown in Figure~\ref{fig:architecture}. As we can see, our network consists of several parts: Token Pyramid Module, Semantics Extractor, Semantics Injection Module and Segmentation Head. The Token Pyramid Module takes an image as input and produces the token pyramid. The Vision Transformer is used as a semantics extractor, which takes the token pyramid as input and produces scale-aware semantics. The semantics are injected into the tokens of the corresponding scale for augmenting the representation by the Semantics Injection Module. Finally, Segmentation Head uses the augmented token pyramid to perform the segmentation task. 
Next, we present 
the details of these modules.

\subsection{Token Pyramid Module}
\label{sec:buaa}
Inspired by MobileNets~\cite{sandler2018mobilenetv2}, the proposed Token Pyramid Module consists of stacked MobileNet blocks~\cite{sandler2018mobilenetv2}. Different from MobileNets, the Token Pyramid Module does not aim to obtain rich semantics and large receptive field, but
uses fewer blocks to build a token pyramid. 
We show the layer settings of the Token Pyramid Module 
in Subsection~\ref{sec:variants}.

As shown in Figure~\ref{fig:architecture},
taking an image $\mathbf{I} \in \mathbb{R}^{3 \times H \times W}$ as input,
where $3, H, W$ indicate the RGB channels, height, width of $\mathbf{I}$ respectively,
our Token Pyramid Module first passes the image through some MobileNetV2 blocks~\cite{sandler2018mobilenetv2} to produce a series of tokens $\{\mathbf{T}^{1}, ..., \mathbf{T}^{N}\}$,
where $N$ indicates the number of scales%
\footnote{We say that the tokens from different stages belong to different scales.}.
Afterwards, the tokens $\{\mathbf{T}^{1}, ..., \mathbf{T}^{N}\}$, are average pooled to the target size, \textit{e.g.}, 
$\mathbb{R}^{\frac{H}{64} \times \frac{W}{64}}$.
Finally, the tokens from different scales are concatenated along the channel dimension to produce the new tokens. The new tokens will be fed into the Vision Transformer to produce Scale-aware Semantics. Because the new tokens are 
of 
small quantity, the Vision Transformer 
can 
run with very low computation cost even if the new tokens have large channels.

\subsection{Vision Transformer as Scale-aware Semantics Extractor}
The Scale-aware Semantics Extractor consists of 
a few 
stacked Transformer blocks. The number of Transformer blocks is $L$. The Transformer block consists of the multi-head Attention module, the Feed-Forward Network (FFN) and residual connections. To keep the spatial shape of tokens and reduce the numbers of reshape, we replace the Linear layers with a $1 \times 1$ convolution layer. Besides, all of TopFormer’s non-linear activations are ReLU6~\cite{howard2017mobilenets} instead of GELU function in ViT. 

For the Multi-head Attention module, we follow the settings of LeViT~\cite{graham2021levit}, and set the head dimension of keys $K$ and queries $Q$ to have $D = 16$, the head of values $V$ to have $2D=32$ channels.  Decreasing the channels of $K$ and $Q$ will reduce computational cost when calculating attention maps and output. Meanwhile, we also drop the Layer normalization layer and append a batch normalization to each convolution. The batch normalization can be merged with the preceding convolution during inference, which 
can 
run faster over layer normalization.

For the Feed-Forward Network, we follow~\cite{yuan2021incorporating,huang2021shuffle} to enhance the local connections of Vision Transformer by inserting a depth-wise convolution layer between the two $1 \times 1$ convolution layers. The expansion factor of FFN is set to
2 to reduce the computational cost. The number of Transformer blocks is $L$ and then the number of heads will be given in subsection~\ref{sec:variants}.

As shown in Figure~\ref{fig:architecture}, the Vision Transformer takes the tokens from different scales as input. To further reduce the computation, the average pooling operator is used to reduce the numbers of tokens from different scales to  $\frac{1}{64\times64}$ of the input size. The pooled tokens from different scales have the same resolution, and they are concatenated together as the input of the Vision Transformer. The Vision Transformer 
can 
obtain full-image receptive field and rich semantics. To be more specific, the global self-attention exchanges information among tokens 
along the 
spatial dimension. The $1 \times 1$ convolution layer will exchange information among tokens from different scales. In each Transformer block, the residual mapping is learned after exchanging information of tokens from all scales, then residual mapping is added into tokens to augment the representation and semantics. Finally, the scale-aware semantics are obtained after passing through several transformer blocks.

\subsection{Semantics Injection Module and  Segmentation Head}
After obtaining the scale-aware semantics, we 
%
%
add them with the other tokens $T^{N}$ directly. However, there is a 
significant 
semantic gap between the tokens $\{\mathbf{T}^{1}, ..., \mathbf{T}^{N}\}$ and the scale-aware semantics. To this end, the Semantics Injection Module is introduced to alleviate the semantic gap before fusing these tokens. As shown in Fig.~\ref{fig:architecture}, the Semantics Injection Module (SIM) takes the local tokens of the  Token Pyramid module and the global semantics of the Vision Transformer as input. The local tokens are
passed through the $1 \times 1$ convolution layer, followed by a batch normalization to produce the feature to be injected. The global semantics are fed into the $1 \times 1$ convolution layer followed by a batch normalization layer and a sigmoid layer to produce semantics weights, meanwhile, the global semantics also passed through the $1 \times 1$ convolution layer followed by a batch normalization. The three outputs have the same size. Then, the global semantics are injected into the local tokens by Hadamard production and the global semantics are also added with the feature after the injection. The outputs of the several SIMs share the same number of channels, denoted as $M$.

After the semantics injection, the augmented tokens from different scales capture both rich spatial and semantic information, which is critical for semantic segmentation. Besides, the semantics injection alleviates the semantic gap among tokens. The proposed Segmentation head firstly up-samples the low-resolution tokens to the same size as the high-resolution tokens and element-wise sums up the tokens from all scales. Finally, the feature is passed through two convolutional layers to produce the final segmentation map.

\begin{table*}[thbp]
    \centering
    \small
    \begin{tabular}{llcccc}
    \toprule
    Method  & Encoder & Params(M) & FLOPs(G) & mIoU & Latency(ms) \\
    \midrule
    FCN-8s~\cite{long2015fully}  & MobileNetV2  & $9.8$  & $39.6$  & $19.7$ & 1015 \\
    PSPNet~\cite{zhao2017pyramid} & MobileNetV2 & $13.7$ & $52.2$ & $29.6$ & 1065\\
    
    DeepLabV3+~\cite{chen2018encoder}  & MobileNetV2 & $15.4$ & $25.8$ & $38.1$ & 1035\\
    DeepLabV3+~\cite{chen2018encoder} & EfficientNet & $17.1$ & $26.9$ & $36.2$ & 970 \\
    DeepLabV3+~\cite{chen2018encoder} & ShuffleNetV2-1.5x & $16.9$ & $15.3$ & $37.6$ & 960\\
    
    Lite-ASPP~\cite{chen2018encoder} & ResNet18 & $12.5$ & $19.2$ & $37.5$ & 648\\
    Lite-ASPP~\cite{chen2018encoder} & MobileNetV2 & $2.9$ & $4.4$ & $36.6$ & 235\\
    
    R-ASPP~\cite{sandler2018mobilenetv2} & MobileNetV2 & $2.2$ & $2.8$ & $32.0$ & 177\\
    LR-ASPP~\cite{howard2019searching} & MobileNetV3-Large & $3.2$ & $2.0$ & $33.1$ & 126\\
    LR-ASPP~\cite{howard2019searching} & MobileNetV3-Large-reduce & $1.6$ & $1.3$ & $32.3$ & 81 \\
    HRNet-W18-Small~\cite{YuanCW19} & HRNet-W18-Small & $4.0$ & $10.2$ & $33.4$ & 639\\
    HR-NAS-A~\cite{ding2021hrnas} & Searched & $2.5$ & $1.4$ & $33.2$ & -\\
    HR-NAS-B~\cite{ding2021hrnas} & Searched & $3.9$ & $2.2$ & $34.9$ & -\\
    Segformer~\cite{xie2021segformer} & MiT-B0 & $3.8$ & $8.4$ & $37.4$ & 770\\
    Semantic FPN~\cite{kirillov2019panoptic} & ConvMLP-S & $12.8$ & $33.8$ & $35.8$ & 777\\
    \midrule
    Ours & TopFormer-T$^\dagger$ & $1.4$ & $0.5$ & $32.5$ & 32\\
    Ours & TopFormer-T & $1.4$ & $0.6$ & $32.8$ & 43\\
    Ours & TopFormer-S & $3.1$ & $1.2$ & $36.1$ & 74\\
    Ours & TopFormer-B & $5.1$ & $1.8$ & $37.8$ & 110\\
    \bottomrule
    \end{tabular}
    \vspace{8pt}
    \caption{Results on ADE20K \textit{val} set. Latency and GFLOPs calculation adopt images with $512\times512$ resolution as input. $^\dagger$ indicates results are obtained with $448\times448$ resolution as input. Latency is measured based on a single Qualcomm Snapdragon 865 processor. All results are evaluated with single thread.  Following the settings in MMSegmentation, batch size=32 is used for the CNN-based methods. For a fair comparison with Segformer, the batch size=16 is used for TopFormer. The mIoU is reported with single-scale inference.} 
    \label{tab:main_results_ade20k}
    \vspace{-5mm}
\end{table*}

\subsection{Architecture and Variants} \label{sec:variants}

To customize the network of various complexities, we introduce TopFormer-Tiny (TopFormer-T), TopFormer-Small (TopFormer-S) and TopFormer-Base (TopFormer-B), respectively. 

The model size and FLOPs of the base, small and tiny models are given in the Table~\ref{tab:main_results_ade20k}. The base, small and tiny models have 8, 6 and 4 heads in each multi-head self-attention module, respectively, and have $M=256$, $M=192$ and $M=128$ as the target numbers of channels. For more details of network configures, please refer to the supplementary materials.

 
To achieve better trade-offs between accuracy and the actual latency, we choose the tokens from last three scales $T^2,T^3$ and $T^4$ as the inputs of SIM and the segmentation head.

\section{Experiments} \label{sec:exp}

In this section, we first conduct experiments on 
several public 
datasets. We describe implementation details and compare results with other works 
for 
semantic segmentation tasks. We then conduct ablation studies to analyze the effectiveness and efficiency of different parts. Finally, we report the performance on object detection task to show the generalization ability of our method.

\subsection{Semantic Segmentation}

\subsubsection{Datasets}
We perform experiments over three datasets, ADE20K~\cite{zhou2017scene}, PASCAL Context~\cite{mottaghi2014role} and COCO-Stuff~\cite{caesar2018coco}. The mean of class-wise intersection over union (mIoU) is set as our evaluation metric. The full-precision TopFormer models are converted to TNN~\cite{tnn2019}, the latency is then measured on an ARM-based computing core. \textbf{ADE20K}: The ADE20K dataset contains 25K images in total, covering 150 categories. All images are split into 20K/2K/3K for training, validation, and testing. \textbf{PASCAL Context}: The Pascal Context dataset has 4998 scene images for training and 5105 images for testing. There are 59 semantic labels and 1 background label. \textbf{COCO-Stuff}: The COCO-Stuff~\cite{caesar2018coco} dataset augments COCO dataset with pixel-level stuff annotations. There are 10000 complex images selected from COCO. The training set and test set consist of 9K and 1K images respectively.

\subsubsection{Implementation Details}

Our implementation is 
built upon 
MMSegmentation~\cite{mmseg2020} and Pytorch.
It utilizes ImageNet pre-trained TopFormer\footnote{The TopFormer-base achieve 75.3\% Top-1 acc. on ImageNet-1K.} as the backbone.
The standard BatchNorm~\cite{ioffe2015batch} layer is replaced by the Synchronize BatchNorm to collect the mean and standard-deviation of BatchNorm across multiple GPUs during training. 
For ADE20K dataset, we follow Segformer to use 160K scheduler and batch size is 16. The training iteration of COCO-Stuff and PASCAL Context is 80K.
For all methods and datasets, the initial learning rate is set as 0.00012 and weight decay is 0.01. A ``poly'' learning rate scheduled with factor 1.0 is used. On ADE20K, we adopt the same data augmentation strategy as ~\cite{xie2021segformer} for fair comparison. The training images are augmented by first randomly scaling and then randomly cropping out the fixed size patches from the resulting images. In addition, we also apply random resize, random horizontal flipping, random cropping etc. On COCO-Stuff and PASCAL Context, we use the default augmentation strategy of ~\cite{mmseg2020}. We resize and crop the images to $480\times480$ for PASCAL Context and $512\times512$ for COCO-Stuff during training. Finally, we report the single scale results on validation set to compare with other methods. During inference, we follow the common strategy to rescale the short side of images to training cropping size for ADE20K and COCO-Stuff. As for PASCAL Context, the images are resized to $480\times480$ and then fed into our network.

\subsubsection{Experiments on ADE20K}
We compare our TopFormer with the previous approaches on the ADE20K validation set in Table~\ref{tab:main_results_ade20k}. Actual latency is measured on the mobile device with  a single Qualcomm Snapdragon 865 processor.
Here, we choose light-weight vision transformers (ViT)~\cite{ding2021hrnas,xie2021segformer,kirillov2019panoptic,ma2018shufflenet} and efficient convolution neural networks (CNNs)~\cite{sandler2018mobilenetv2,howard2019searching,sun2019deep,he2016deep} as the encoder. Besides, the various decoders are also included in Table~\ref{tab:main_results_ade20k}. 
Among all methods in Table~\ref{tab:main_results_ade20k}, Deeplabv3+ based on MobilenetV2 achieve best mIoU (38.1\%), however, the latency is more than 1000 ms, which restrict its application on mobile devices.

Among these CNNs based baselines, the approach which takes mobilenetV3-large~\cite{howard2019searching} as encoder and LR-ASPP as decoder, achieves good trade-off between computation (2.0 GFLOPs) and accuracy (33.1 mIoU). Following~\cite{howard2019searching}, we also reduce the channel counts of all feature
layers in the last stage for further reducing the computation cost, denoted as MobileNetV3-Large-reduce. Based on the lighter backbone, LR-ASPP could achieve 32.3\%  mIoU with the lower latency~(81 ms). Our small version of TopFormer is 3.8\% more accurate compared to a LR-ASPP model with comparable latency. The tiny version of TopFormer could achieve comparable performance compared to LR-ASPP with $2\times$ less computation (0.6G vs.\  1.3G). Lite-ASPP~\cite{chen2018encoder} is the reduced channel version of Deeplabv3+.

\begin{table} \setlength{\tabcolsep}{4pt}
    \centering
    \small
    \begin{tabular}{ccccc}
    \toprule
    Input  & Top-1 acc.  & mIoU  & params(M) \\
    \midrule
    Token Pyramid  & 75.3 & 37.8 & 5.1   \\
    Token of the last scale & 74.8 & 36.6 & 5.2   \\
    \bottomrule
    \end{tabular}
    \caption{Ablation studies on Token Pyramid as input.}
    \label{tab:token_pyramid_input}
    \vspace{-4mm}
\end{table}

\begin{table} \setlength{\tabcolsep}{6pt}
    \centering
    \small
    \begin{tabular}{cccc}
    \toprule
    Output  & mIoU  & params(M) & FLOPs(G) \\
    \midrule
    $\{\frac{1}{4},\frac{1}{8},\frac{1}{16},\frac{1}{32}\}$ & 38.3 & 5.1 & 3.32  \\
    $\{\frac{1}{8},\frac{1}{16},\frac{1}{32}\}$ & 37.8 & 5.1 & 1.82 \\
    $\{\frac{1}{16},\frac{1}{32}\}$ & 36.4 & 5.0 & 1.41 \\
    \bottomrule
    \end{tabular}
    \caption{Ablation studies on Token Pyramid as output.}
    \label{tab:token_pyramid_out}
    \vspace{-4mm}
\end{table}

Among these ViT based baselines, HR-NAS-B~\cite{ding2021hrnas} uses search techniques to introduce Transformer block into the HRNet~\cite{sun2019deep} design, also achieves good trade-off between computation amount (2.2 GFLOPs) and accuracy (34.9 mIoU). Our small version of TopFormer is 1.2\% more accurate compared to HR-NAS-B model with fewer computation. SegFormer~\cite{xie2021segformer} achieves great performance (37.4 mIoU) with fewer parameters (3.8M), although SegFormer adopts the efficient multi-head self-attention, the computation is still heavy due to a large number of tokens. Our base version of TopFormer could achieve comparable performance compared to SegFormer with more than $4\times$ less computation (1.8 GFLOPs vs.\  8.4 GFLOPs).

To achieve real-time segmentation on the ARM-based mobile device, we resize the input image to $448 \times 448$ and feed it into TopFormer-tiny, the inference time is reduced to 32ms with a slight performance drop. To the best of our knowledge, it is the first ViT based method could achieve real-time segmentation on the ARM-based mobile device with competitive results.

\begin{table} \setlength{\tabcolsep}{8pt}
    \centering
    \small
    \begin{tabular}{lcccc}
    \toprule
    Model & Params(M) & FLOPs(M) & mIoU\\ 
        \midrule
        Baseline    & 0.3   & 573   & 22.5 \\
        +SASE       & 1.4   & 643   & 32.8 \\
        +FFNs       & 1.4   & 642   & 30.4 \\
        \midrule
        +PSP        & 1.7   & 654   & 27.4 \\
        +ASPP       & 2.3   & 698   & 28.2 \\
    \bottomrule
    \end{tabular}
    \caption{Ablation studies on Scale-aware Semantics Extractor.}
    \label{tab:sase}
    \vspace{-4mm}
\end{table}

\subsubsection{Ablation Study}
We first conduct ablation experiments to discuss the influence of different components,
including the token pyramid, Semantic Injection Module and Segmentation Head.
Without loss of generality, all results are obtained by training on the training set and evaluation on the validation (val) set.

\paragraph{The influence of the Token Pyramid.}
Here, we discuss the Token Pyramid from two aspects, the influence of taking Token Pyramid as input and the influence of choosing Tokens from different scales as output.
As 
reported 
in Table~\ref{tab:token_pyramid_input}, we conduct the experiments that take stacked tokens from different scales as input of the Semantics Extractor, and take the last token as input of the Semantics Extractor, respectively. For fair comparison, we append a $1\times1$ convolution layer to expand the channels as same as the stacked tokens. The experimental results demonstrate the effectiveness of using the token pyramid as input.

After obtaining scale-aware semantics, the SIM will inject the semantics into the local Tokens. To pursuit better trade-off between the accuracy and the computation cost, we try to choose tokens from different scales for injecting. As shown in Table~\ref{tab:token_pyramid_out}, using tokens from $\{\frac{1}{4},\frac{1}{8},\frac{1}{16},\frac{1}{32}\}$ could achieve the best performance with the heaviest computation. Using tokens from $\{\frac{1}{16},\frac{1}{32}\}$ achieves 
worse performance with the lightest computation. To achieve a good trade-off between the accuracy and the computation cost, we choose to use the tokens from $\{\frac{1}{8},\frac{1}{16},\frac{1}{32}\}$ in all other experiments.


\begin{table}
    \centering
    \small
    \begin{tabular}{cccc}
    \toprule
    SigmoidAttn         & SemInfo   & FLOPs(G) & mIoU \\
    \midrule
    \checkmark &                & 1.809 & 37.3\\
                & \checkmark    & 1.809 & 37.0\\
     \checkmark &\checkmark     & 1.820 & 37.8\\
    \bottomrule
    \end{tabular}
    \caption{Ablation studies on Semantic Injection Module.}
    \label{tab:sim}
    \vspace{-4mm}
\end{table}

\paragraph{The influence of the Scale-aware Semantics Extractor.}
Here, we have conducted experiments on Topformer-T to check the SASE. The results are shown in the Table~\ref{tab:sase}. Here, we use Topformer without SASE as the baseline. Adding SASE brings about 10\% mIoU gains,  which is a significant improvement. To verify the multi-head self-attention module (MHSA) in the Transformer block, we remove all MHSA modules and add more FFNs for a fair comparison. The results demonstrate the MHSA could bring about 2.4\% mIoU gains, which is an efficient and effective module under the careful architecture design. Meanwhile, we compare the SASE with the popular contextual models, such as ASPP and PPM, on the top of TPM. As shown in  Table~\ref{tab:sase}, ``+SASE'' could achieve better performance with much less computation cost than ``+PSP'' and ``+ASPP''. The experimental results demonstrate that the SASE is more appropriate for use in mobile devices.

\begin{figure}[!t]
    \centering
    \includegraphics[width=0.95\linewidth]{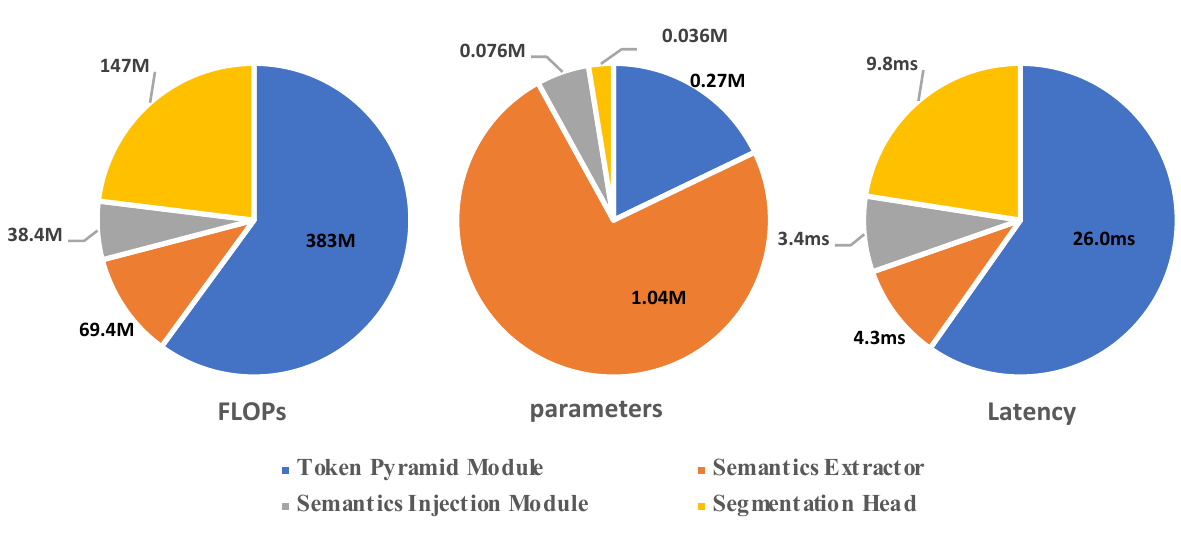}
    \caption{The statistics of the computation, parameters, and latency. Latency and FLOPs are measured with input resolution $512\times512$.} 
    \label{fig:tiny_stats} 
    \vspace{-4mm}
\end{figure}

\paragraph{The influence of the Semantic Injection Module and Segmentation Head.}
Due to the close relationship of Semantic Injection Module and Segmentation Head, we discuss these two together.
Here, we discuss the design of Semantic Injection Module at first. As shown in Table~\ref{tab:sim}, multiplying the local tokens and the semantics after a Sigmoid layer, denoted as ``SigmoidAttn''. Adding the semantics from Semantics Extractor into the corresponding local tokens, denoted as ``SemInfo''. Compared with ``SigmoidAttn'' and ``SemInfo'', adding ``SigmoidAttn'' and ``SemInfo'' simultaneously could bring pretty improvement with a little extra computation.

\begin{figure}[!t]
    \centering
    \includegraphics[width=1\linewidth]{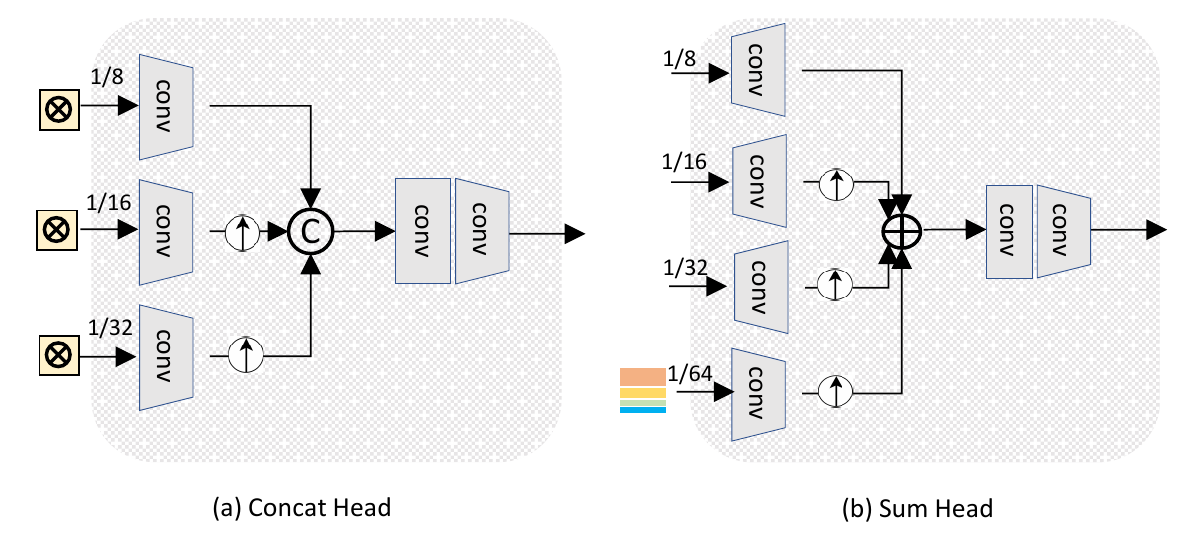}
    \caption{Different segmentation heads in Token Pyramid Transformer. The Batch-norm layer and activation layer are omitted in the Figure. } 
    \label{fig:seg_head} 
    \vspace{-3mm}
\end{figure}

Here, we discuss the design of Segmentation Head. After passing the feature into Semantic Injection Module, the output hierarchical features are with both strong semantics and rich spatial details. The proposed segmentation head simply adds them together and then uses two $1\times1$ convolution layers to predict the segmentation map. We also design the other two Segmentation Heads, as shown in Figure~\ref{fig:seg_head}. The ``Sum Head'' is identical to only adding ``SemInfo'' in SIM. The ``Concat Head'' uses a $1\times1$ convolution layer to reduce the channels of the  outputs of SIM, then the features are concatenated together. Compared with ``Concat Head'' and ``Sum Head'', the current segmentation head could achieve better performance. 

\begin{table} \setlength{\tabcolsep}{8pt}
    \centering
    \small
    \begin{tabular}{lcc}
    \toprule
    Setting         & Params(M)         & mIoU \\
    \midrule
    Concat Head     & $5.044$             & $37.1$ \\
    Sum Head        & $4.978$             & $37.0$ \\
    Ours            & $5.063$             & $37.8$ \\
    \bottomrule
    \end{tabular}
    \caption{Ablation studies on Segmentation Head.}
    \label{tab:seg_head}
    \vspace{-6mm}
\end{table}

\paragraph{The influence of the width in SIM.} In the paper, we donate the number of channels in SIM as $M$. Here, we study the influence of different M in SIM and find a suitable $M$ to achieve a good trade-off. As shown in Table~\ref{tab:sim_m}, the $M=256,192,128$ achieve similar performance with very close computation. Thus, we set $M=128,192,256$ in tiny, small and base model, respectively.
\paragraph{The influence of output stride.} The tokens from different stages are pooled to fixed resolution, namely the output stride. The results of different resolutions are shown in Table~\ref{tab:stride_setting}. s32, s64, s128 are denoted that the resolution of pooled tokens are $\frac{1}{32\times32},\frac{1}{64\times64},\frac{1}{128\times128}$ of input size. Considering the trade-off of computation and accuracy, we choose s64 as the output stride of input tokens of the Semantics Extractor.
\begin{table}[!h] 
\setlength{\tabcolsep}{8pt}
    \centering
    \small
    \begin{tabular}{cccc}
    \toprule
    $M$  & mIoU    & FLOPs(G) & Params(M)\\
    \midrule
    $128$             & $37.3$ & $1.5$ & $4.9$   \\
    $192$             & $37.4$ & $1.6$ & $5.0$  \\
    $256$             & $37.8$ & $1.8$ & $5.1$  \\
    \bottomrule
    \end{tabular}
    \caption{Ablation studies on different \#channels in SIM.}
    \label{tab:sim_m}
    \vspace{-5mm}
\end{table}

\begin{table} \setlength{\tabcolsep}{6pt}
    \centering
    \small
    \begin{tabular}{lcccc}
    \toprule
    Stride  & mIoU  & FLOPs(G) & Latency(ms) \\
    \midrule
    s32  & $38.8$ & $2.6$ & $159$ \\
    s64 & $37.8$ & $1.8$ &  $110$ \\
    s128 & $35.7$ & $1.6$ & $100$  \\
    \bottomrule
    \end{tabular}
    \caption{Ablation studies on output strides.}
    \label{tab:stride_setting}
    \vspace{-5mm}
\end{table}

\begin{table}
    \centering
    \small
    \resizebox{\columnwidth}{!}{
    \begin{tabular}{llccll}
    \toprule
    \multirow{2}*{Methods}   &
    \multirow{2}*{Backbone}   &
    \multicolumn{2}{c}{FLOPs(G)}    & \multirow{2}*{mIoU$^{60}$}  &
    \multirow{2}*{mIoU$^{59}$} \\
    
    &   & Backbone      & Head  &   & \\
    \midrule
    DeepLabV3+      & MBV2-s16   & $2.46$        & $19.78$  & $38.59$     & $42.34$ \\
    DeepLabV3+      & ENet-s16  & $3.22$        & $19.78$  & $39.19$    & $43.07$ \\
    LR-ASPP         & MBV3-s16   & $1.67$        & $0.37$  & $35.05$    & $38.02$ \\
     \midrule
    Ours            & TopFormer-T       & $0.44$       & $0.09$   & $36.41$       & $40.39$ \\
    Ours            & TopFormer-S       & $0.80$       & $0.18$   & $39.06$       & $43.68$ \\
    Ours            & TopFormer-B       & $1.25$       & $0.29$   & $41.01$       & $45.28$ \\
    \bottomrule
    \end{tabular}}
    \caption{Results on Pascal Context \textit{test} set.}
    \label{tab:pc}
    \vspace{-6mm}
\end{table}
\paragraph{The statistics of the computation, parameters and latency.}
Here, we make the statistics of the computation, parameters and latency of the proposed TopFormer-Tiny. As shown in Figure~\ref{fig:tiny_stats}, although the Semantics Extractor has most parameters (74\%), the FLOPs and actual latency of the Semantics Extractor is relatively low (about 10\%).

\subsubsection{Experiments on Pascal Context}
We compare our TopFormer with the previous approaches on the Pascal Context test set in Table~\ref{tab:pc}. We evaluate the performance over 59 categories and 60 categories (including background), respectively. It is obvious that our approach achieves better performance than all previous approaches based on CNNs or ViT with fewer computation. For better understanding, the FLOPs of the backbone and the head are measured, respectively. The proposed method 
can 
achieve best performance with the lightest backbone and head.

\subsubsection{Experiments on COCO-Stuff}
We compare our TopFormer with the previous approaches on the COCO-Stuff validation set in Table~\ref{tab:stuff}. The FLOPs of the backbone and the head are measured, respectively. It can be seen that our approach achieves the best performance, and the base version of TopFormer is 8\% more accurate compared to a MobileNetV3 model with comparable computation.


\begin{table}
    \centering
    \small
    \resizebox{\columnwidth}{!}{
    \begin{tabular}{llccll}
    \toprule
    \multirow{2}*{Methods}   &
    \multirow{2}*{Backbone}   &
    \multicolumn{2}{c}{FLOPs(G)}    & \multirow{2}*{mIoU} \\
    
    &   & Backbone      & Head  & \\
    \midrule
    PSPNet          & MobileNetV2-s8       & $7.84$        & $45.10$       & $30.14$ \\
    DeepLabV3+      & MobileNetV2-s16      & $2.46$        & $23.44$       & $29.88$ \\
    DeepLabV3+      & EfficientNet-s16      & $3.66$        & $23.44$       & $31.45$   \\
    LR-ASPP         & MobileNetV3-s16      & $1.89$        & $0.48$        & $25.16$     \\
    \midrule
    Ours            & TopFormer-T   & $0.49$        & $0.15$        & $28.34$ \\
    Ours            & TopFormer-S   & $0.89$        & $0.29$        & $30.83$ \\
    Ours            & TopFormer-B   & $1.38$        & $0.45$        & $33.43$ \\
    \bottomrule
    \end{tabular}}
    \caption{Results on COCO-Stuff \textit{test} set.}
    \label{tab:stuff}
    \vspace{-5mm}
\end{table}

\begin{table}
    \centering
    \small
    \resizebox{\columnwidth}{!}{
    \begin{tabular}{llcccc}
    \toprule
    \multirow{2}*{Backbone}   &
    \multirow{2}*{mAP}   &
    \multicolumn{2}{c}{FLOPs(G)}    & 
    \multicolumn{2}{c}{Params(M)} \\
    
    &   & Backbone      & Overall  & Backbone      & Overall \\
    \midrule
    ShuffleNetv2  & 25.9 & 2.6 & 161 & 0.8 & 10.4 \\
    TopFormer-T  & 27.1 & 2.1 & 160 & 1.0 & 10.5 \\
    \midrule
    MobileNetV3   & 27.2 & 4.7 & 162 & 2.8 & 12.3  \\
    TopFormer-S  & 30.4 & 3.7 & 162 & 2.9 & 12.3 \\
    TopFormer-B  & 31.6 & 5.5 & 163 & 4.8 & 14.0 \\
    \bottomrule
    \end{tabular}}
    \caption{COCO object detecion based on RetinaNet.}
    \label{tab:detection_retinanet}
    \vspace{-6mm}
\end{table}

\subsection{Object Detection}
To further demonstrate the generalization ability of the proposed TopFormer, we conduct object detection task on COCO dataset. COCO consists of 118K images for training, 5K for validation and 20K for testing. We train all models on train2017 split and evaluate all methods on val2017 set. We choose RetinaNet~\cite{lin2017focal} as object detection methods and adopt different backbones to produce feature pyramid. Our implementation is 
built 
on MMdetection~\cite{MMDetection} and Pytorch. For the proposed TopFormer, we replace the segmentation head with the detection head in RetinaNet. 
As shown in Table~\ref{tab:detection_retinanet}, The RetinaNet based on TopFormer could achieve better performance than MobileNetV3 and ShuffleNet with lower computation.

\section{Conclusion and Limitations} \label{sec:Conclusion}
In this paper, we present a new architecture for mobile vision tasks. With a combination of the advantages of CNNs and ViT, the proposed TopFormer achieves a good trade-off between accuracy and the computational cost. The tiny version of TopFormer could yield real-time inference on an ARM-based mobile device with competitive result. The experimental results demonstrate the effectiveness of the proposed method. The major limitation of TopFormer is the minor improvements on object detection. We will continue to promote the performance of object detection. Besides, we will explore the application of TopFormer in dense prediction in the future work.

\section*{Acknowledgement} This work was in part supported by NSFC (No.\ 61733007, No.\  61876212, No. 62071127 and No.\  61773176) and the Zhejiang Laboratory Grant (No.\  2019NB0AB02 and No.\  2021KH0AB05).



{\small
\bibliographystyle{ieee_fullname}
\bibliography{egbib}
}

\clearpage

This chapter presents additional materials and results. We give the ImageNet pretraining results in Section ~\ref{sec:imagenet_cls}. Then we describe the specific network structure in Section~\ref{sec:net_stru}. Next, we give the performance on Cityscapes. Finally, some visual results are provided.

\section{ImageNet Pre-training} \label{sec:imagenet_cls}
\begin{figure}[!t]
    \centering
    \includegraphics[width=1\linewidth]{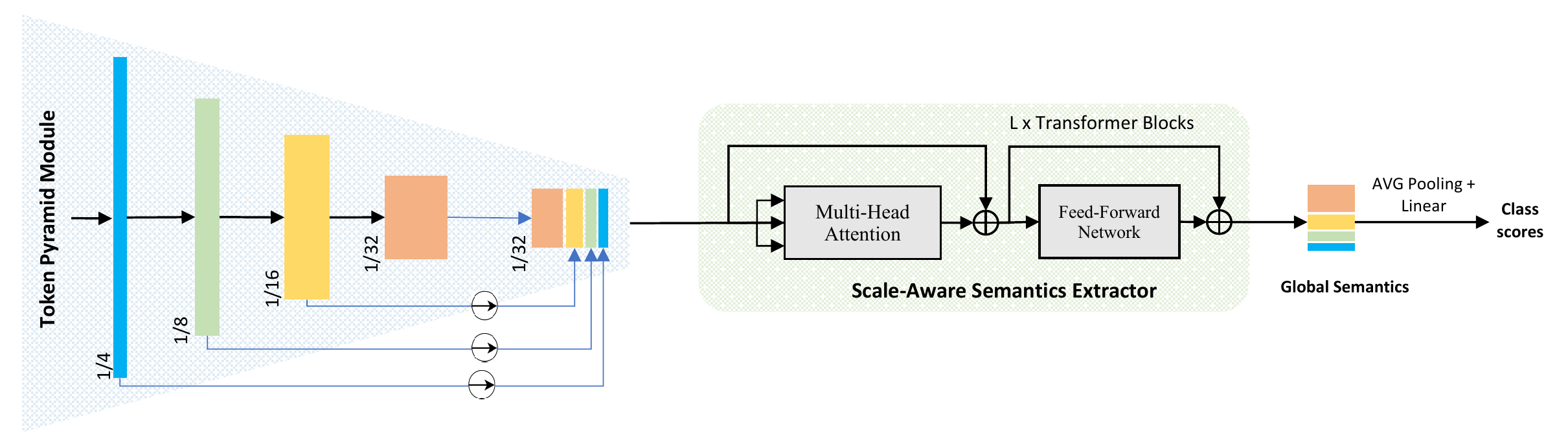}
    \caption{The classification architecture of the proposed Token Pyramid Transformer. } 
    \label{fig:cls_arch} 
    \vspace{-0mm}
\end{figure}

For fair comparison, we also use the ImageNet pre-trained parameters as initialization. As shown in Figure~\ref{fig:cls_arch}, the classification architecture of the proposed TopFormer appends the average pooling layer and Linear layer on the global semantics for producing class scores. Due to the small resolution ($224\times224$) of input images, we set the target resolution of input tokens of the Semantics Extractor is $\frac{1}{32\times32}$ of input size. The classification results are shown in Table~\ref{tab:cls_res}. Because our target task is mobile semantic segmentation, we do not explore more technologies, e.g. more epochs and distillation uesd in LeViT, to further improve the accuracy. In the future work, we will continue to improve the classification accuracy.

\begin{table}[!h]
    \centering
    \small
    \begin{tabular}[h]{lccc}
    \toprule
    Model       & Params    & FLOPs(M)     & Top-1 Acc(\%)\\
    \midrule
    TopFormer-T & $1.50$M      & $126$         & $66.2$   \\
    TopFormer-S & $3.11$M      & $235$         & $72.3$   \\
    TopFormer-B & $5.07$M      & $373$         & $75.3$   \\
    \bottomrule
    \end{tabular}
    \caption{The results on ImageNet classification.}
    \label{tab:cls_res}
    \vspace{-2mm}
\end{table}

\section{Network Structure} \label{sec:net_stru}
The detailed network structures are given in Table~\ref{tab:detail_architecture}. 
Although the Token Pyramid Module have the most layers, as the statistics of the computation and parameters in the paper, the ViT-based Semantics Extractor accounts for the vast majority of parameters.

\section{The Performance on Cityscapes} \label{sec:city_res}
\paragraph{Training Settings} Our implementation is based on MMSegmentation and Pytorch. We perform 80K iterations. The initial learning rate is 0.0003 and weight decay is 0.01. A poly learning rare scheduled with factor 1.0 is used. For full-resolution version, the training images are randomly scaling and then cropping to fixed size of $1024\times1024$. As for the half-resolution version, the training images are resized to $1024\times512$ and randomly scaling, the crop size is $1024\times512$. We follow the data augmentation strategy of Segformer for fair comparison. 

\paragraph{Experimental results} To validate the performance of the proposed method, we directly fed a full-resolution input and a half-resolution input into the trained segmentation models for testing, respectively. As shown in Table~\ref{tab:city_res}, the proposed method with a full-resolution, denoted as Ours(f), achieves about 2.6\% higher accuracy in mIoU than L-ASPP based on MobileNetV2 with lower computation. The experimental results demonstrate that TopFormer could achieve good trade-off between accuracy and computation even if the input image is with large resolution.

\subsection{Visualization}
\begin{figure*}[!t]
    \centering
    \includegraphics[width=1.0\linewidth]{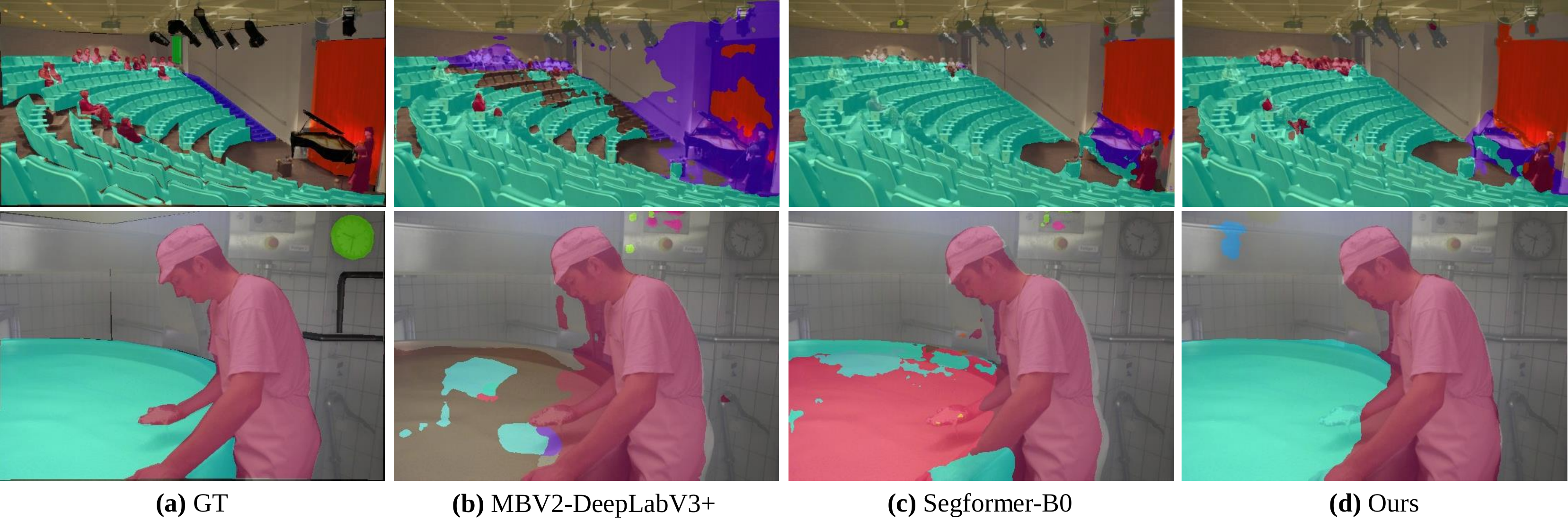}
    \caption{The visualization of the Ground Truth, MBV2-Deeplabv3+, SegFormer-B0 and the proposed TopFormer on ADE20K val set. We use TopFormer-B to conduct visualization.} 
    \label{fig:vis} 
    \vspace{-1mm}
\end{figure*}

We present some visualization comparisons among the proposed TopFormer and other CNNs- and ViT-based methods on the ADE20K validation (val) set in Figure~\ref{fig:vis}. Here, we choose deeplabv3+ based on mobilenetV2 as a representative of CNNs-based methods and Segformer as a representative of ViT-based methods. These two methods both have larger model size and computational cost. As shown in Figure~\ref{fig:vis}, the proposed method could achieve better segmentation results than these two methods.

\begin{table}[!t]  \setlength{\tabcolsep}{5pt}
    \centering
    \small
    \begin{tabular}{lccc}
    \toprule
    Methods  & Encoder  & GFLOPs & mIoU \\
    \midrule
    FCN      & MobileNetV2  & 317.1 & 61.5 \\
    PSPNet      & MobileNetV2   & 423.4 & 70.2  \\
    Segformer   & MiT-B0        & 17.7 & 71.9 \\
    L-ASPP   & MobileNetV2        & 12.6 & 72.7 \\
    LR-ASPP   & MobileNetV3-large        & 9.7 & 72.4 \\
    LR-ASPP   & MobileNetV3-small        & 2.9 & 68.4 \\
    \midrule
    Ours(h) & TopFormer-B  & $2.7$ & $70.7$   \\
    Ours(f) & TopFormer-B  & $11.2$ & $75.0$   \\
    \bottomrule
    \end{tabular}
    \caption{Results on Cityscapes val set. Ours(f) and Ours(h) are denoted as taking a full-resolution input (i.e., $1024 \times 2048$) and a half-resolution input (i.e., $512 \times 1024$). }
    \label{tab:city_res}
    \vspace{-0mm}
\end{table}

\begin{table*}[!t]
    \renewcommand{\arraystretch}{1.3}
    \centering \small
    \begin{tabular}{ c | c | c  c  c  c  }
        \toprule[-0.2em]
        \textbf{Stage} & \textbf{Output size}  & \textbf{Tiny} & \textbf{Small} & \textbf{Base} \\ 
        \bottomrule[0.1em]
        \multirow{10}{*}{Token Pyramid Module} & \multirow{2}{*}{$256\times256$}  & \multicolumn{3}{c}{Conv,$3\times3$, 16, 2} \\ 
        &              & \multicolumn{3}{c}{MB, 3, 1, 16, 1} \\ \cline{2-5}
        & \multirow{2}{*}{$128\times128$}   &  {MB, 3, 4, 16, 2} & {MB, 3, 4, 24, 2} & {MB, 3, 4, 32, 2} \\
        &      &  {MB, 3, 3, 16, 1} & {MB, 3, 3, 24, 1} & {MB, 3, 3, 32, 1} \\ \cline{2-5}
        & \multirow{2}{*}{$64\times64$}     &  {MB, 5, 3, 32, 2} & {MB, 5, 3, 48, 2} & {MB, 5, 3, 64, 2} \\
        &     &  {MB, 5, 3, 32, 1} & {MB, 5, 3, 48, 1} & {MB, 5, 3, 64, 1} \\ \cline{2-5}
        & \multirow{2}{*}{$32\times32$}     &  {MB, 3, 3, 64, 2} & {MB, 3, 3, 96, 2} & {MB, 3, 3, 128, 2} \\
        &     &  {MB, 3, 3, 64, 1} & {MB, 3, 3, 96, 1} & {MB, 3, 3, 128, 1} \\ \cline{2-5}
        & \multirow{3}{*}{$16\times16$}     &  {MB, 5, 6, 96, 2} & {MB, 5, 6, 128, 2} & {MB, 5, 6, 160, 2} \\
        &     &  {MB, 5, 6, 96, 1} & {MB, 5, 6, 128, 1} & {MB, 5, 6, 160, 1} \\ 
        &     &                    & {MB, 3, 6, 128, 1} & {MB, 3, 6, 160, 1} \\ \hline
        
        Semantics Extractor  & $8\times8$ & L=4,H=4 & L=4,H=6 & L=4,H=8   \\ \hline
        Semantics Injection Module  & $16\times 16$,$32\times32$,$64\times64$ &  M=128 & M=192 & M=256 \\ 
        \bottomrule[0.1em]
        FLOPs  &  &  0.6G & 1.2G & 1.8G \\ 
        \bottomrule[0.1em]
    \end{tabular}
    \caption{Detailed architecture configs of the proposed method. The input is with resolution $512\times512$. For Token Pyramid Module, Conv refers to regular convolution layer, [MB, 5, 3, 48, 1] refers to MobileNetV2 block with kernel size=5, expand ratio=3, output channels=48 and stride=1.  For Semantics Extractor, L is the number of Transformer Blocks. H is the number of heads in a multi-head self-attention block.
    }
    \label{tab:detail_architecture}
    \vspace{-0mm}
\end{table*}

\end{document}